\documentclass{article}
\usepackage{spconf,amsmath,graphicx}
\usepackage{caption}
\usepackage{booktabs}
\usepackage{array}
\usepackage{makecell}
\usepackage{multirow} 
\usepackage{verbatim}
\usepackage{subfig}
\usepackage{amssymb} 
\usepackage{hyperref}

\title{Multimodal Transformer Using Cross-Channel attention for \\Object Detection in Remote Sensing Images}
\name{Bissmella Bahaduri, Zuheng Ming, Fangchen Feng, Anissa Mokraoui }
\address{L2TI Laboratory, University Sorbonne Paris Nord, Villetaneuse, France}
\begin{document}
%
\maketitle

\begin{abstract}
Object detection in Remote Sensing Images (RSI) is a critical task for numerous applications in Earth Observation (EO). Differing from object detection in natural images, object detection in remote sensing images faces challenges of scarcity of annotated data and the presence of small objects represented by only a few pixels. Multi-modal fusion has been determined to enhance the accuracy by fusing data from multiple modalities such as RGB, infrared (IR), lidar, and synthetic aperture radar (SAR). To this end, the fusion of representations at the mid or late stage, produced by parallel subnetworks, is dominant, with the disadvantages of increasing computational complexity in the order of the number of modalities and the creation of additional engineering obstacles. Using the cross-attention mechanism, we propose a novel multi-modal fusion strategy for mapping relationships between different channels at the early stage, enabling the construction of a coherent input by aligning the different modalities. By addressing fusion in the early stage, as opposed to mid or late-stage methods, our method achieves competitive and even superior performance compared to existing techniques. Additionally, we enhance the SWIN transformer by integrating convolution layers into the feed-forward of non-shifting blocks. This augmentation strengthens the model's capacity to merge separated windows through local attention, thereby improving small object detection. Extensive experiments prove the effectiveness of the proposed multimodal fusion module and the architecture, demonstrating their applicability to object detection in multimodal aerial imagery. Our code is available
at \href{https://github.com/Bissmella/Small-object-detection-transformers.git}{here}.
\end{abstract}
\section{Introduction}
\label{sec:intro}
Object detection in Remote Sensing Images (RSI) including aerial images is a critical task enabling the identification and localization of objects within satellite or aerial imagery. It has numerous applications for Earth Observation (EO) such as environmental monitoring, climate change, urban planning, and military surveillance~\cite{ahmad2021hyperspectral}. Given the availability of different sensors onboard satellites and UAVs, multimodal fusion has been an active area of interest for research. Given the smaller object size and limitation of annotated data availability~\cite{sharma2020yolors}, multimodal data has been used to improve the performance of diverse computer vision tasks in RSI including object detection. Previous work has shown that using neural networks for fusing multiple modalities has remarkable benefits for tasks such as semantic segmentation~\cite{park2017rdfnet}, video description~\cite{nagrani2021attention}, and action recognition~\cite{song2020modality}.

Multimodal fusion methods can be classified according to the stage at which the fusion is performed: early-stage fusion, mid-stage fusion, and late-stage fusion~\cite{feng2020deep}. Early-stage fusion methods typically employ simple techniques like concatenation or averaging to integrate multimodal data at the early stage of the model. Although early-stage fusions are usually simple and easy to use, their performance lags behind more sophisticated mid- or late-stage methods. In mid-stage and late-stage fusion, modality features are extracted by independent feature extractors sub-networks, often referred to as streams in the literature, before being fused. The key distinction between mid-stage and late-stage fusion lies in their subsequent processing steps: in mid-stage fusion, the fused features are further processed by a main network, whereas in late-stage fusion, the feature extractor streams constitute the entirety of the network.

Depending on the fusion technique employed, multimodal fusion approaches can also be categorized into aggregation-based, alignment-based, and attention-based methods. Aggregation-based methods~\cite{hazirbas2017fusenet} rely on straightforward operations like concatenation and averaging to fuse modalities, a process that can occur at any stage—early, mid, or late. Alignment-based methods~\cite{song2020modality} leverage regularization loss to align embeddings from different modalities, typically executed at the late stage of fusion. The more recent attention-based methods~\cite{mohla2020fusatnet, nagrani2021attention} utilize attention mechanisms for fusion, commonly occurring at mid or late stages. A detailed review of some related attention-based fusion methods will be presented in the next section.

\begin{figure*}[hbt!]
    \centering
    \includegraphics[width=0.92\textwidth]{ 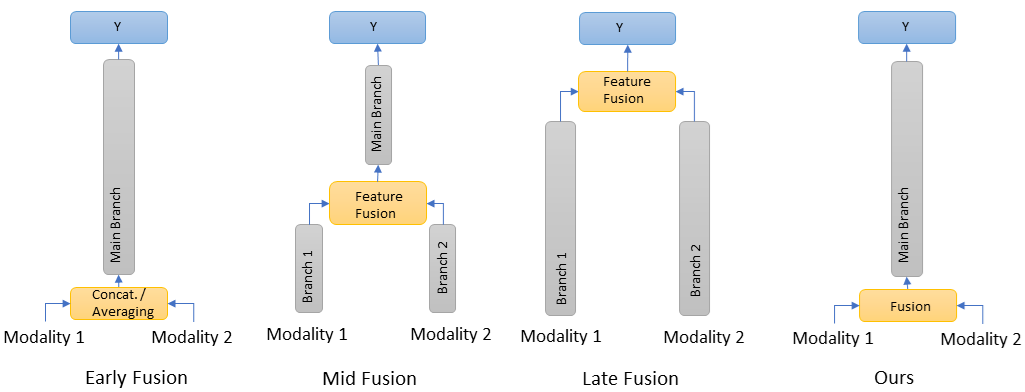}
    \caption{An illustration of different fusion strategies. In contrast to mid-stage and late-stage fusions which involve fusion at the feature level extracted by distinct parallel subnetworks, our approach targets fusion at the early stage.}
    \label{fig1:comparison}
\end{figure*}

Despite the achievements and good results of these methods, several challenges persist. While aggregation-based methods offer simplicity, they often overlook potential modal misalignments and intra-modality dependencies. On the other hand, alignment-based methods, while aimed at minimizing alignment loss, may not adequately address intramodal information exchanges. Moreover, regardless of the fusion approach, the use of subnetworks for feature extraction from different modalities remains widespread. These subnetworks are trained to extract modal features independently, with fusion occurring in subsequent steps, as shown in Fig~\ref{fig1:comparison}. This family of methods increases computational complexity and adds additional technical hurdles, such as determining the optimal step and scale at which fusion is performed. This challenge is particularly pronounced when the independent parallel subnetworks are composed of recently developed Vision Transformers (ViT)~\cite{dosovitskiy2020image}, where computational costs increase quadratically. In addressing this issue, some works choose CNN for modality-specific streams and propose to use transformers for the multimodal fusion module~\cite{mohla2020fusatnet}. This integration of CNNs and ViTs also occurs for object detection where existing fusion methods are mainly CNN-based~\cite{sharma2020yolors, mohla2020fusatnet}.

\begin{figure}[hbt!]
    \centering
    \includegraphics[width=0.4\textwidth]{ 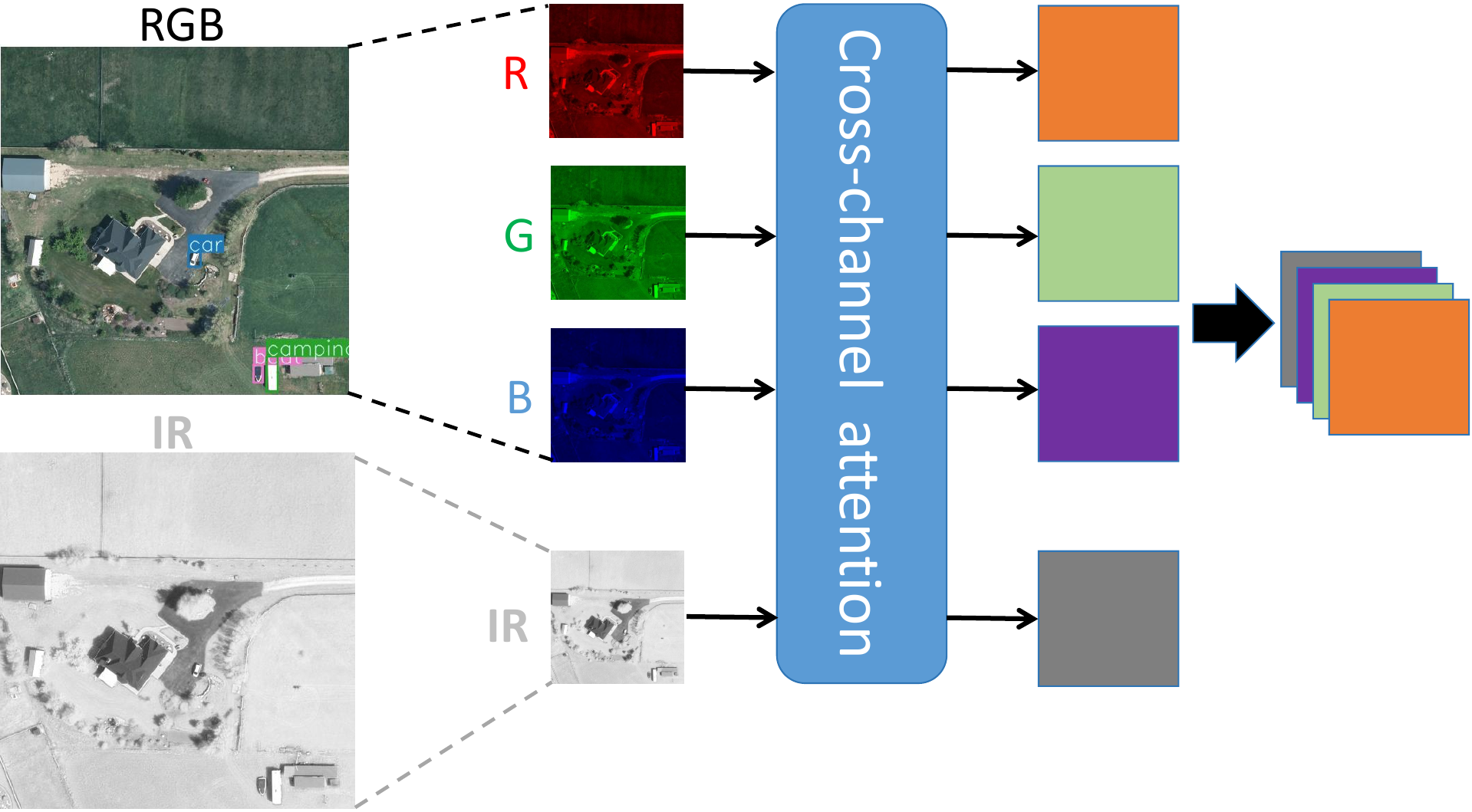}
    \caption{Combining multimodal inputs using cross-channel attention instead of simple channel-wise concatenation.}
    \label{fig1:cc-attention}
\end{figure}

In this paper, we consider object detection in remote sensing images with the presence of two modalities, RGB and infrared (IR). We propose an early-stage multimodal cross-attention fusion method based on the cross-attention mechanism (Fig.~\ref{fig1:cc-attention}). We show that different modalities can be fused at the early stage without requiring additional independent subnetworks for each modality. The key concept behind the proposed method involves processing the three RGB channels individually instead of considering them as a single modality. Notably, the absence of modality-specific subnetworks enables scalability to accommodate any number of modalities without a substantial increase in model size. Additionally, this strategy proves effective in eliminating the need for additional subnetworks dedicated to modality feature extraction. We demonstrate that the cross-channel attention fusion can outperform or match the performance of state-of-the-art mid-stage fusion based on the CNN backbone.

Additionally, we propose to use SWIN~\cite{liu2021Swin} as the main backbone which uses the shifted window mechanism to enable the merging of neighboring patches separated by local window attention. Given that this shifting mechanism is only operative in half of the blocks, we propose augmenting the Feed-Forward network (FFN) with a convolutional layer while maintaining the same embedding dimension. This improvement aims to strengthen the network's ability to capture local information and facilitate the integration of neighboring patches in different windows. Hereafter, we refer to these blocks as convolutional-shifting blocks. Furthermore, we show the necessity of this backbone improvement for the proposed fusion strategy and we believe that this is due to the rich information present in the fusion module output.

To summarize, the main contributions of this work are: 
1) We introduce a new cross-channel attention module that allows for the early alignment of different modalities.

\noindent2) We propose a convolutional-shifting window that incorporates convolutional layers in FFN to assist in merging separated windows in local attention enhancing the detection performance.

\noindent3) The extensive experiments demonstrate the superiority of the proposed approach, highlighting its applicability for object detection in multimodal aerial imagery.



\section{Related Work}
\label{sec:relatedwork}

\noindent\textbf{\textit{Attention based multimodal fusion:}}
Recently attention-based multimodel fusion has caught attention. In contrast to aggregation-based approaches, which rely on simple concatenation, attention-based fusion methods employ self and cross-attention mechanisms to fuse modality features at the mid or late-stage. Cross-modal attention is used to enrich the main modality with lidar information in~\cite{mohla2020fusatnet}, while~\cite{sun2021deep} employs attention at various stages between two separate branches for multi-modal fusion. Additionally, attention is utilized in~\cite{nagrani2021attention} for the mid-stage fusion of image and audio modalities through the introduction of extra bottleneck tokens. Similarly,~\cite{chen2023incomplete} introduces additional learnable fusion tokens and aggregates modalities for token fusion using Bi-LSTM.

Compared to aggregation-based methods, existing attention-based approaches replace concatenation with more sophisticated feature-level attention mechanisms. However, they still require feature extractors for each modality. The difference between our proposed cross-channel attention fusion and existing attention-based methods is that, in our proposed method, fusion occurs before any substantial processing and featurization.

 \noindent\textbf{\textit{Multimodal fusion for object detection:}} Over the past decade, deep learning methods have been widely explored for multimodal data fusion to improve object detection performance. \cite{hazirbas2017fusenet} is an aggregation-based method and its performance is affected by the scale and stage at which aggregation occurs, which is a common problem for mid-stage fusion methods. \cite{bousmalis2016domain} is an alignment-based method that aligns multimodal features based on similarity regulation. One of the limitations of this method is that it ignores modal-specific information.


In terms of architectural design, a notable commonality among these existing methods is their reliance on subnetwork branches for feature extraction. Indeed, \cite{liu1611multispectral} explored early, middle, and late-stage fusion using CNNs for object detection and claimed that mid-stage fusion yielded the optimal performance. The following works all followed this principle. \cite{fang2022cross} uses the transformer self-attention mechanism for fusion at multiple stages between two modality branches. Manish et al.~\cite{sharma2020yolors} proposed a real-time framework for object detection in multi-modal remote sensing imagery by performing a mid-stage fusion of RGB and IR modalities. More recently, Zhang et al.~\cite{zhang2023superyolo} proposed the CNN-based SuperYOLO and studied various fusion strategies, including early-stage concatenation. It's worth noting that their proposed multimodal fusion also relies on subnetworks comprising squeeze and excitation blocks~\cite{hu2018squeeze} maintaining consistent input sizes.
Different from these methods, we propose cross-channel attention fusion without the need for separate subnetworks. Furthermore, whereas the aforementioned methods consider RGB as a unified modality, we opt to treat each color channel independently. This approach enables us to leverage not only the cross-information between two modalities (RGB and IR) but also among all three RGB channels.

\noindent\textbf{\textit{Window-based Vision Transformers/CNN}}. 
The landscape of computer vision has undergone a substantial transformation with the emergence of ViT~\cite{dosovitskiy2020image}, showcasing advancements across a broad spectrum of visual tasks. 
However, while ViTs exhibit commendable performance, they come with certain limitations, notably a substantial computational burden and lack of sense of locality. To address these limitations, \cite{liu2021Swin} enhanced the vanilla ViTs by introducing hierarchical architectures and localized windows. These enhancements have found practical applications in single-modal aerial image object detection~\cite{cao2023Swin}. Drawing from this body of work, more recent approaches endeavor to combine the strengths of CNNs with ViTs~\cite{wu2021cvt}. This combination capitalizes on the respective advantages of both CNNs and ViTs, thus offering promising prospects for computer vision applications such as classification~\cite{wu2021cvt}, and Face Presentation Attack Detection (PAD)~\cite{ming2022vitranspad}. In this paper, we utilize SWIN architecture as the main backbone, with an enhancement on joining the separated windows by incorporating a convolutional layer in the feed-forward network of non-shifting blocks.

\section{Proposed method}
\label{sec:multiTrans}
\begin{figure*}[!ht]
\begin{minipage}[t]{0.62\textwidth} 
    \centering
    \includegraphics[width=1\textwidth]{ 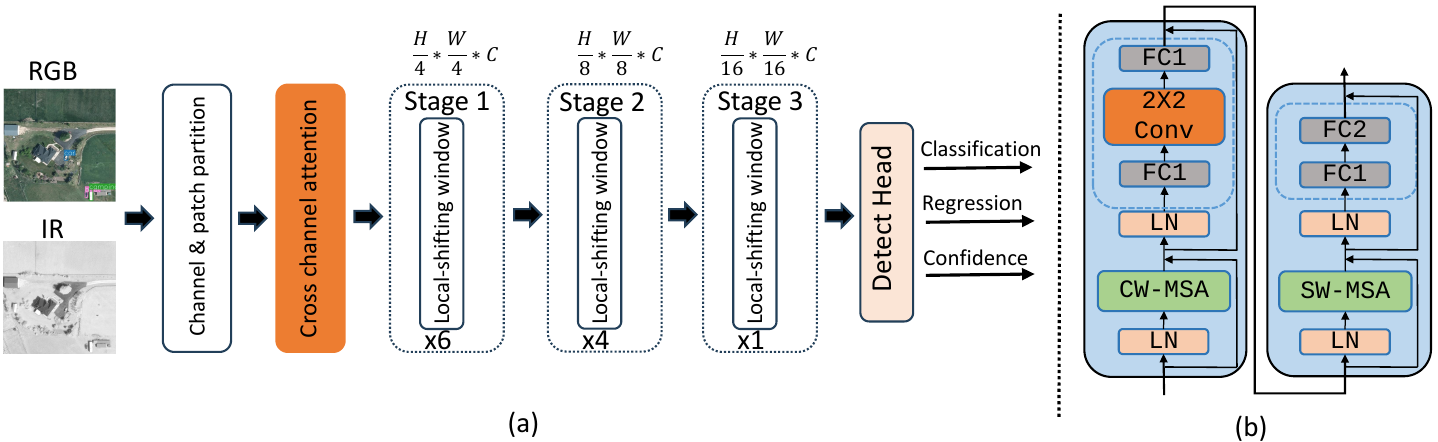}
    \caption{(a) The overall architecture based on Swin-like backbone for multimodal object detection in RSI;
(b) Convolutional-shifting window module.}
    \label{fig:gen-arch}
\end{minipage}
\vspace{0.1pt}
\vline
\vspace{0.2pt}
\begin{minipage}[t]{0.34\textwidth} 
    \centering
    \includegraphics[width=1\textwidth]{ 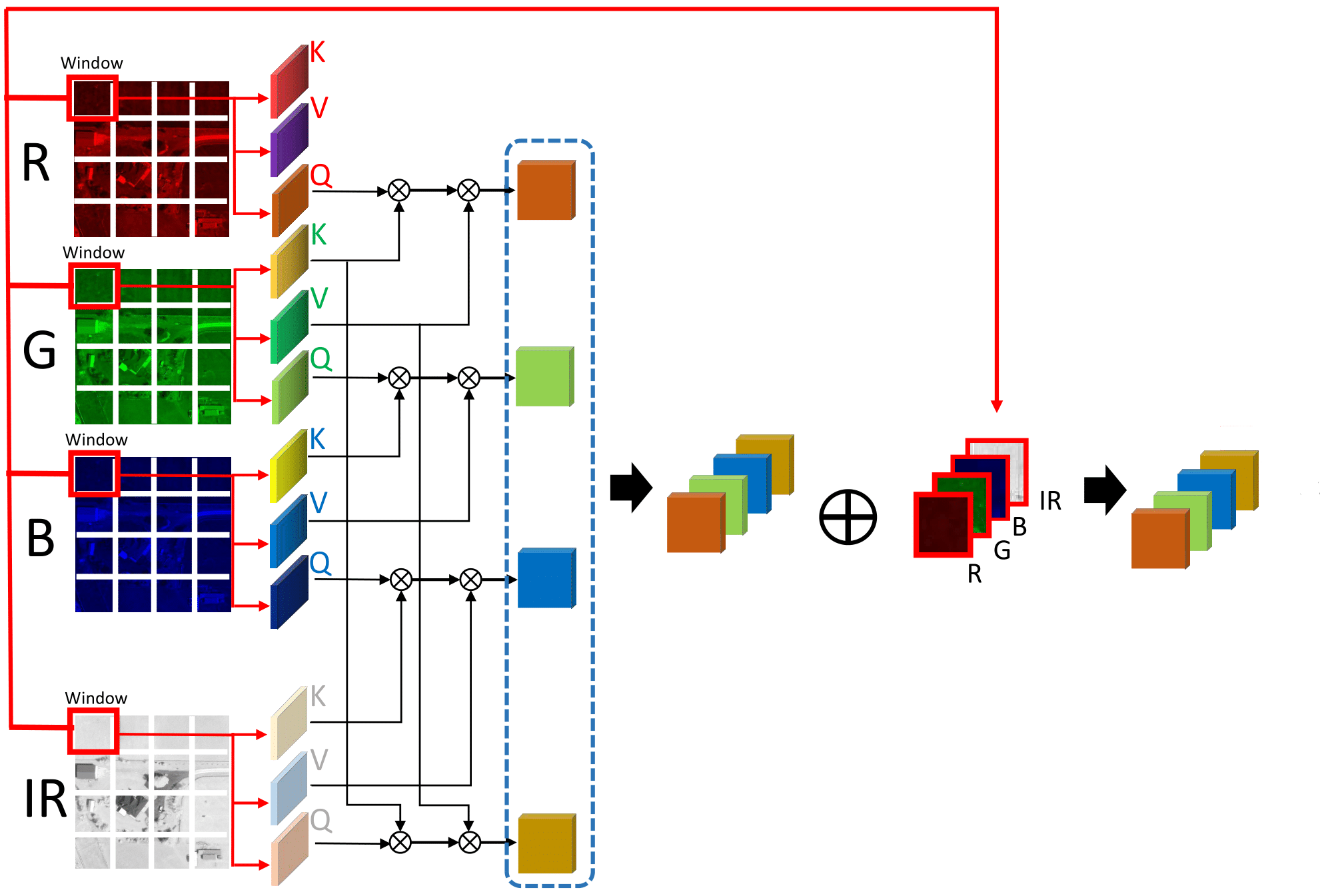}
    \caption{Cross-channel attention module for RGB and IR images multimodal fusion.}
    \label{fig:cc-attention}
\end{minipage}
\end{figure*}

\subsection{Overall architecture}
Illustrated in Fig~\ref{fig:gen-arch}, the proposed architecture is composed of three components: the cross-channel attention fusion module, a feature extraction backbone featuring SWIN-like blocks enhanced with our proposed convolutional-shifting window mechanism, and a YOLO-based detection head as employed in~\cite{zhang2023superyolo}. The choice of this detection head is motivated by the simplicity inherent in the one-step detection method, widely adopted in the field, facilitating comparisons.

\subsection{Cross-channel attention for multimodal fusion}
\label{sec:crossCA}



We present our proposed cross-channel attention fusion module in Fig~\ref{fig:cc-attention}. Considering $X_{\text{RGB}} \in \mathbb{R}^{H \times W \times 3}$, and $X_{\text{IR}} \in \mathbb{R}^{H \times W}$ as RGB and IR modality image inputs respectively, following the standard practice in ViT methodologies, each channel (R, G, B, and IR) undergoes tokenization and patch splitting. The resulting patched channels can be denoted respectively as $X_{\text{r}}, X_{\text{g}}, X_{\text{b}}, X_{\text{ir}} \in \mathbb{R}^{\frac{H}{s} \times \frac{W}{s} \times d}$ where $s$ represents the patch size and $d$ denotes the embedding dimension. Subsequently, the obtained $X_{\text{r}}, X_{\text{g}}, X_{\text{b}}, X_{\text{ir}}$ are fused based on the proposed cross-channel attention fusion module which relies on the cross-attention mechanism. Mathematically, the attention mechanism can be generally formulated as in equation~\eqref{eq:cc-attention}:
\begin{gather}
\label{eq:cc-attention}
\text{Attention}(Q, K, V) = \text{softmax}\left(\frac{Q \cdot K^T }{\sqrt{d}}\right) \cdot V
\end{gather}
where $Q, K, V$ are the query, key, and value matrices and $d$ is the embedding dimension. Cross-attention and self-attention represent two applications of the attention mechanism. Cross-attention involves the asymmetric combination of two embedding sequences, where one sequence acts as a query input, while the other serves as key and value inputs, thereby capturing the interdependence between the two sequences. We demonstrate the effectiveness of cross-attention for fusion by contrasting it with a fusion technique based on self-attention. Detailed comparisons can be found in the experimental section.

In our fusion process, we propose to apply cross-attention individually to each channel (R, G, B, IR). Specifically, each channel acts as a query, and one of the remaining channels is chosen to fulfill both key and value roles in the cross-attention mechanism. We select the complementary channel to minimize its spectral proximity to the query channel. The formed four pairs are (R, G), (G, B), (B, IR), and (IR, G). The fused output comprises the concatenation of all these fused channels along with a residual branch. To streamline computational complexity, we omit the feed-forward network from the cross-channel attention, as its role in altering the channel dimension is redundant within this context.

We highlight that, in our cross-channel attention framework, we treat each channel as an independent modality, thus incorporating the relationship between channels alongside information exchange with the IR modality. This decomposition of RGB enables distinct fusion mechanisms for each RGB channel and the IR modality, recognizing potential variations in their relationships. To illustrate this, we also tested a cross-attention-based fusion technique treating RGB as a single modality. Detailed comparisons are provided in the experimental section. Moreover, we deliberately align the number of output channels of the fusion module with the input channels. This choice ensures direct or indirect linkage among the four channels. Additionally, it substantially reduces computational complexity.

\subsection{Convolutional-shifting window-based backbone}

We base our backbone on SWIN which adopts a hierarchical approach with multiple stages. SWIN divides the input image into non-overlapping patches at multiple scales, enabling more efficient processing of both local and global information. The hierarchical transformer architecture allows SWIN to capture long-range dependencies while maintaining computational efficiency. In our SWIN-based backbone, three stages process the image at different scales starting from a higher resolution scale to low resolution in the third stage. Since the objects in RSI are often small and densely packed into a few pixels, we choose a higher number of blocks in the initial stage where the resolution remains high, while progressively reducing the number of blocks in later stages decreasing the resolution by a factor of 2. This backbone enables us to learn hierarchical multi-resolution features in a fine-to-coarse manner with a higher focus in the first stage for better detection of small objects (see Fig~\ref{fig:gen-arch} (a)).


In addition, a well-recognized limitation of window-based Vision Transformers is their segregation of neighboring patches across different windows. To address this challenge, the SWIN Transformer introduces a shifting window mechanism, albeit restricted to only half of its blocks in a way that each nonshifting block is followed by a shifting window block. In our approach, we seek to enhance connectivity across all blocks and imbue the architecture with a heightened sense of locality. To achieve this, we introduce an extra convolutional layer with a kernel size of two by two positioned between two Fully Connected (FC) layers within the FFN  while keeping the embedding dimension fixed (the orange block as shown in Fig~\ref{fig:gen-arch} (b)). The use of a two-by-two kernel size and a fixed dimension is intended to prevent an increase in the computational complexity of the model. This augmentation not only promotes greater coherence but also enhances the network's perception of spatial proximity. 

The enhanced blocks can be utilized across all three stages of the process. Through empirical investigation, we have determined that the optimal performance is attained when the enhanced blocks are deployed in stages 1 and 2. A comparison of various combinations is provided in the experimental section for further insight.

\section{Experiments}
\label{sec:majhead}
\begin{table*}[ht!]
    \centering	
    \begin{tabular}{c|c|c|c|c|c|c|c|c|c}    	
        \midrule[1pt]
         \textbf{Method}&\textbf{Car}&\textbf{Pickup}&\textbf{Camping}&\textbf{Truck}&\textbf{Other}&\textbf{Tractor}&\textbf{Boat}&\textbf{Van}&$\textbf{mAP}_\textbf{50}$  \\
         \hline
         YOLOv3~\cite{redmon2018yolov3}&84.57&72.68&67.13&61.96&43.04&65.24&37.10&58.29&61.26   \\
         
         YOLOv4~\cite{bochkovskiy2020yolov4}&85.46&72.84&72.38&62.82&48.94&68.99&34.28&54.66&62.55\\
         YOLOv5~\cite{ultralytics_yolov5}&84.33&72.95&70.09&61.15&49.94&67.35&38.71&56.65&62.65\\

         YOLOrs~\cite{sharma2020yolors}&84.15&78.27&68.81&52.60&46.75&67.88&21.47&57.91&59.73\\
         YOLO-Fine~\cite{pham2020yolo}&79.68&74.49&77.09&\textbf{80.97}&37.33&70.65&\textbf{60.84}&63.56&68.83\\
         SuperYOLO~\cite{zhang2023superyolo}&89.30&81.48&\textbf{79.22}&67.27&54.29&\textbf{78.88}&55.95&71.41&72.22\\
         \textbf{Ours}&\textbf{89.13}&\textbf{82.70}&76.38&61.57&\textbf{56.32}&77.94&60.36&\textbf{75.84}&\textbf{72.53} \\
         
         \midrule[1pt]
    \end{tabular}
    \caption{Class-wise mean Average Precision $mAP_{50}$ for our proposed method comparing to the state-of-art on VEDAI Dataset.}
    \label{tab:general}
\end{table*}
\subsection{Experimental setup}

We apply our proposed cross-channel attention to the task of object detection in aerial images. We experiment with the VEDAI dataset~\cite{razakarivony2016vehicle}. The dataset contains scenes captured from the same altitude with a resolution of 12.5cm $\times$ 12.5cm per pixel. Each image has 4 channels of RGB and IR. The dataset consists of 1246 images having diverse backgrounds containing highways, pastures, mountains, and urban areas. The images come in both $1024 \times 1024$ and $512 \times 512$ sizes and we have used the $512 \times 512$ images for all the experiments. The task is to identify and localize 11 types of vehicles such as car, pickup, truck, and camping. The dataset is divided into 10 folds for cross-validation evaluation. Following the same protocol as in SuperYOLO~\cite{zhang2023superyolo}, we use the first folder for the ablation studies, and all 10 folders for overall evaluation compared with the state-of-the-art methods. The data is augmented with Hue Saturation Value (HSV), multi-scale, translation, left-right flip, and mosaic augmentation methods similar to SuperYOLO. We considered eight classes in the dataset and ignored classes that have under 50 instances in the dataset.

We have used the standard Stochastic Gradient Descent (SGD)~\cite{robbins1951stochastic} to optimize the network with a momentum of $0.937$ and weight decay of $0.0005$. The models were trained for 300 epochs using one Nvidia A100 GPU. We used the standard detection loss which combines localization, classification, and confidence losses to train the model, and the performances are evaluated using $mAP_{50}$, i.e., detection metric of mean Average Precision at IOU (Intersection over Union) $= 0.5$ for all categories.
\begin{table}[ht]
    \begin{tabular}{l|c|c}
        \midrule[1pt]
         \textbf{Architecture}&\textbf{Fusion Method}&$\textbf{mAP}_\textbf{50}$  \\
         \hline
         \multirow{3}{*}{\makecell{CNN-based 
         \\ (SuperYOLO)}}
         &Pixel fusion& 76.90 \\
         &MF fusion& 77.73 \\
         &\textbf{CC attention}& \textbf{77.9} \\
         \hline
         \multirow{3}{*}{\makecell{ViT-based \\ (Ours)}}
         &IR& 63.79\\
         &RGB& 70.55\\
         &RGB-IR concatenation&74.23\\
         &RGB-IR Vanilla self-attention & 74.03 \\
         &RGB-IR Vanilla cross-attention & 75.64 \\
         &\textbf{RGB-IR CC attention}&\textbf{78.53}\\
         \midrule[1pt]
    \end{tabular}
    \caption{The comparison of the proposed multi-channel attention fusion module with other fusion methods with CNN-based and ViT-based backbones on the VEDAI dataset (Fold-1).}
    \label{tab:cc-attention_effect}
\end{table}

\subsection{Cross-channel attention fusion module}
\label{subsec:ccAttention}
We evaluate the performance of the proposed Cross-Channel (CC) attention fusion module. To illustrate the effectiveness of the cross-channel attention fusion, we designed two fusion techniques based on vanilla self-attention and vanilla cross-attention. For the vanilla self-attention fusion, after the standard ViT patch encoding with separate projections, the obtained embeddings of each modality are concatenated before using a standard self-attention mechanism to perform the fusion. The difference between this configuration and the proposed cross-channel attention is that the self-attention is used instead of the cross-attention and the RGB channels are treated jointly. For the vanilla cross-attention, the cross-attention is used with two modalities (RGB and IR) as separate inputs. Notably, this differs from the proposed cross-channel attention fusion, where attention is applied at the channel level within each modality. All the aforementioned attention mechanisms are employed in a multi-head way.

We tested the proposed fusion module on two types of backbone. We chose the CNN-based SuperYOLO's backbone and our enhanced SWIN-like backbone which is ViT-based. For the CNN-based backbone, we compare the proposed fusion method with pixel-level fusion and feature-level fusion, both studied in SuperYOLO. For the ViT-based backbone, we compare the proposed fusion method with the early-stage concatenation method, vanilla self-attention fusion, and vanilla cross-attention fusion. Performance with a single modality is also presented as a reference. The results are shown in Table~\ref{tab:cc-attention_effect}. 

For the CNN-based backbone, CC-attention fusion outperforms the Pixel-level fusion and the Multimodal Feature-level (MF) fusion used in SuperYOLO. For the ViT-based backbone, the CC-attention fusion outperforms the RGB-IR concatenation by~3.3\% and improves by respectively~15\% and ~8\% compared to the scenario where only IR or RGB images are used. As expected, the proposed cross-channel attention fusion method outperforms the two tested attention-based fusion strategies. It's also interesting to notice that the vanilla cross-attention method outperforms the vanilla self-attention method. These observations validate the advantages of cross-attention and confirm the benefits of treating RGB channels individually. 

\begin{table}[h!]
    \centering
    \begin{tabular}{c|c}
        \midrule[1pt]
         \textbf{Backbone}&$\textbf{mAP}_\textbf{50}$\\
         \hline
         Simple FFN & 71.23 \\
         FFN with Conv2d in stage 1& 75.75 \\
         FFN with Conv2d in stages 1,2 & 76.52 \\
         
         \midrule[1pt]
    \end{tabular}
    \caption{Effect of the convolution in FFN at the 1st and the 2nd stages on the VEDAI dataset (Fold-1).}
    \label{tab:conv2d}
\end{table}

\subsection{Convolutional-shifting window}
We demonstrate the effectiveness of integrating the enhanced convolutional-shifting window across three stages of the backbone. We evaluate three scenarios to assess the impact.
Table~\ref{tab:conv2d} shows the results. We can see that when adding a convolution layer inside the FFN in non-shifting blocks at stage 1, the model outperforms the FFN without convolution by 4.5\% in terms of the $mAP_{50}$. Furthermore, introducing convolution at stages 1 and 2 gains 5.2\% improvement.  
\subsection{Overall performance}
\begin{figure}
    \centering
  \includegraphics[scale=0.5]{ 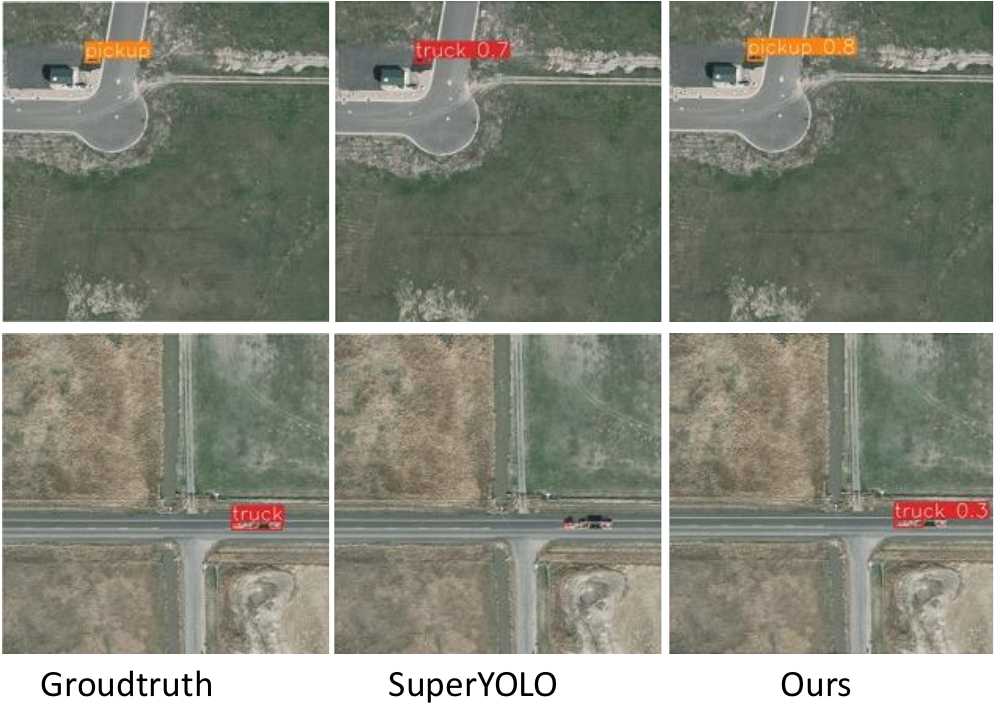}
  \caption{Visual results using our method and SuperYOLO.}
  \label{fig:visualresults}
\end{figure}

The overall performance of our proposed approach is illustrated in Table~\ref{tab:general}. we can see that our proposed method achieves competitive results and outperforms the state-of-the-art CNN model by 0.3\%. Additionally, our method outperforms SuperYOLO in detecting difficult classes with the least number of instances in the training set, namely the Boat, Van, and Other classes. We also illustrate in Fig~\ref{fig:visualresults} a visual comparison of our method and SuperYOLO for two different scenes where only our method has successfully detected and correctly classified the objects. 



\section{Conclusions}
\label{sec:conclusion}
This paper introduces a new cross-channel attention multimodal fusion module that allows for aligning different modalities by learning the relationship between different channels at the early stage. The achieved results demonstrate the module's competitiveness compared to the mid-stage CNN-based fusion methods. Moreover, we propose the convolutional-shifting window, which incorporates convolutional layers in a feedforward network of non-shifting blocks in SWIN. This enhancement facilitates the joining of separated local windows within the SWIN framework, thus augmenting the model's capability to localize and detect small objects. The experiments provide evidence of the competitive and better performance of the proposed approach compared to the mid-stage fusion technique in the context of object detection using multimodal aerial imagery. The results also emphasize the utility of the convolutional-shifting window in enhancing the model's spatial awareness and object detection capabilities.

For future endeavors, exploring the algorithm's applicability across different modalities presents an intriguing avenue. Additionally, investigating pretraining strategies such as self-supervised learning could further enhance the robustness and performance of the method.


\section{Acknowledgements}
\label{sec:acknowledgements}
This work was supported in part by the Laboratoire Commun Intelligence, ReconnaIssance et SurveillancE Réactive (IRISER) LabCom IRISER Project.


\vfill\pagebreak

\bibliographystyle{IEEEbib}

\bibliography{refs}

\end{document}